\documentclass[sigconf]{acmart}
\usepackage{soul}
\usepackage{url}
\usepackage{balance}
\usepackage[utf8]{inputenc}
\usepackage{graphicx}
\usepackage{amsmath}
\usepackage{amsthm}
\usepackage{booktabs}
\usepackage{algorithm}
\usepackage{algorithmic}
\usepackage{subfigure}
\usepackage{multirow}
\usepackage{dsfont}
\usepackage{enumitem}

\usepackage{bm}
\usepackage{stfloats}
\usepackage{colortbl}
\usepackage{newfloat}
\usepackage[labelfont=bf,size=small]{caption}
\definecolor{tab_red}{RGB}{255,174, 185}
\definecolor{tab_yellow}{rgb}{0.99,0.99,0.70}
\definecolor{tab_blue}{RGB}{191,239, 255}
\usepackage[leftcaption]{sidecap}

\PassOptionsToPackage{citecolor=royalblue,breaklinks,colorlinks,pagebackref}{hyperref}

\def\argmax{\operatornamewithlimits{arg\,max}}

\usepackage{newfloat}
\usepackage{listings}
\DeclareCaptionStyle{ruled}{labelfont=normalfont,labelsep=colon,strut=off} 
\floatstyle{ruled}
\newfloat{listing}{tb}{lst}{}
\floatname{listing}{Listing}

\usepackage{pifont}
\definecolor{royalblue}{RGB}{65,105,225} 
\usepackage{xspace}
\makeatletter
\DeclareRobustCommand\onedot{\futurelet\@let@token\@onedot}
\def\@onedot{\ifx\@let@token.\else.\null\fi\xspace}
\def\eg{\emph{e.g}\onedot} 
\def\ie{\emph{i.e}\onedot} 
 
\def\etc{\emph{etc}\onedot}

\makeatother

\AtBeginDocument{%
  }

\begin{document}

\title{ALA: Naturalness-aware Adversarial Lightness Attack}

\author{Yihao Huang}
\orcid{0000-0002-5784-770X}
\affiliation{%
  \institution{Nanyang Technological University}
  \country{Singapore}
}

\author{Liangru Sun}
\affiliation{%
  \institution{East China Normal University}
  \country{China}}

\author{Qing Guo}
\affiliation{%
  \institution{IHPC and CFAR, Agency for Science, Technology and Research}
  \country{Singapore}
}

\author{Felix Juefei-Xu}
\authornote{Work done prior to joining Meta.}
\affiliation{%
 \institution{Meta AI}
 \country{USA}}

\author{Jiayi Zhu}
\affiliation{%
  \institution{East China Normal University}
  \country{China}}

\author{Jincao Feng}
\affiliation{%
  \institution{East China Normal University}
  \country{China}
  }
\affiliation{%
  \institution{Shanghai Industrial Control Safety Innovation Tech. Co., Ltd}
  \country{China}
  }
  
\author{Yang Liu}
\authornote{Yang Liu and Geguang Pu are the corresponding authors.}
\affiliation{%
  \institution{Zhejiang Sci-Tech University}
  \country{China}
  }
\affiliation{%
  \institution{Nanyang Technological University}
  \country{Singapore}
}

\author{Geguang Pu}
\authornotemark[2]
\affiliation{%
  \institution{East China Normal University}
  \country{China}
  }
\affiliation{%
  \institution{Shanghai Industrial Control Safety Innovation Tech. Co., Ltd}
  \country{China}
  }


\renewcommand{\shortauthors}{Yihao Huang et al.}
\renewcommand{\authors}{Yihao Huang, Liangru Sun, Qing Guo, Felix Juefei-Xu, Jiayi Zhu, Jincao Feng, Yang Liu, Geguang Pu}

\begin{abstract}
Most researchers have tried to enhance the robustness of DNNs by revealing and repairing the vulnerability of DNNs with specialized \emph{adversarial examples}. Parts of the attack examples have imperceptible perturbations restricted by $L_p$ norm. However, due to their high-frequency property, the adversarial examples can be defended by denoising methods and are hard to realize in the physical world. To avoid the defects, some works have proposed unrestricted attacks to gain better robustness and practicality. It is disappointing that these examples usually look unnatural and can alert the guards. In this paper, we propose \textbf{A}dversarial \textbf{L}ightness \textbf{A}ttack (ALA), a white-box unrestricted adversarial attack that focuses on modifying the lightness of the images. The shape and color of the samples, which are crucial to human perception, are barely influenced. To obtain adversarial examples with a high attack success rate, we propose unconstrained enhancement in terms of the light and shade relationship in images. To enhance the naturalness of images, we craft the naturalness-aware regularization according to the range and distribution of light. The effectiveness of ALA is verified on two popular datasets for different tasks (\ie, ImageNet for image classification and Places-365 for scene recognition). 
%
\end{abstract}



\keywords{Adversarial Attack; Lightness; Naturalness-aware}


\maketitle

\begin{figure}[t]
    \centering
    \includegraphics[width=\linewidth]{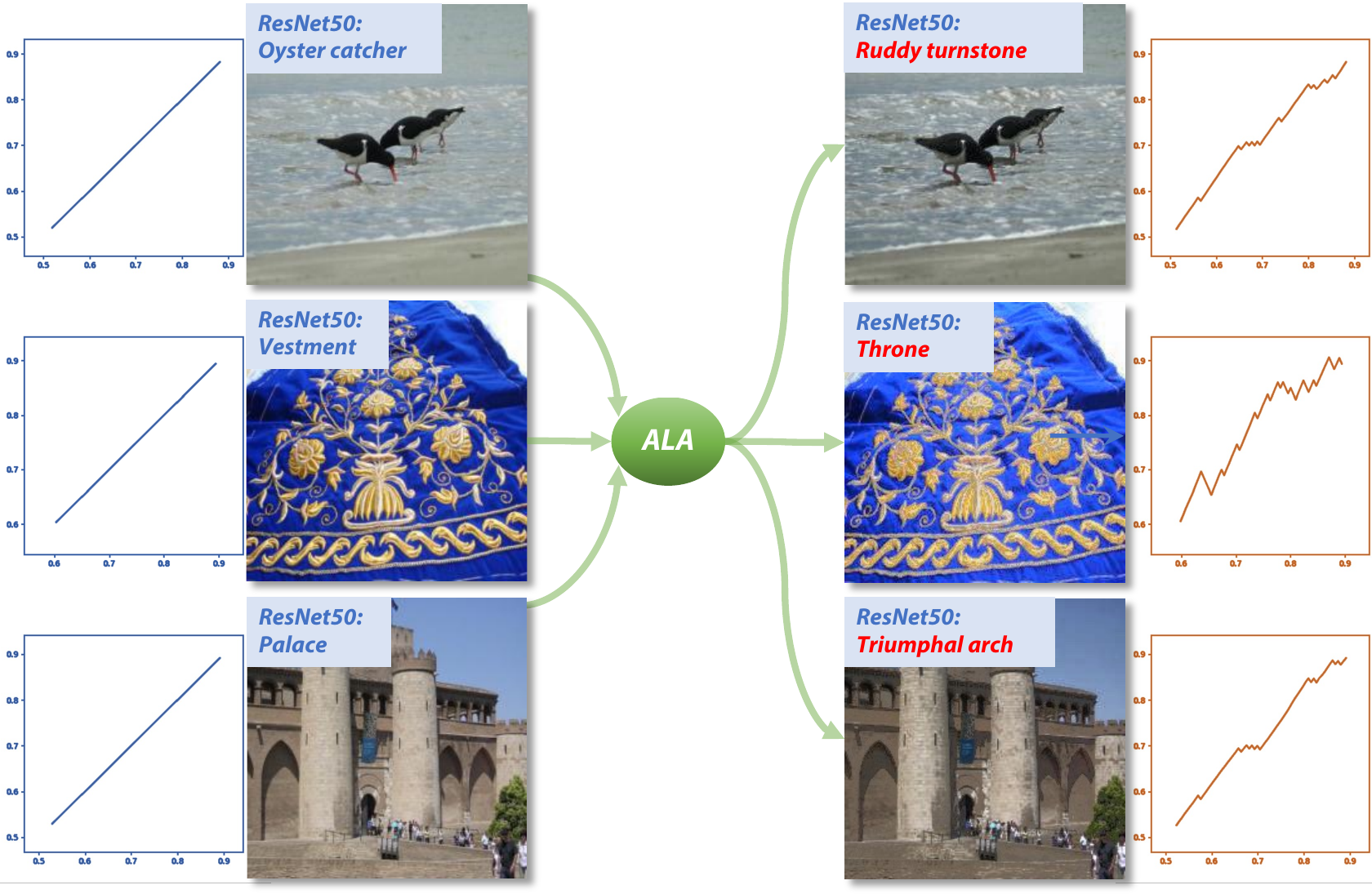}    
    \caption{(L) Original images with their labels (successfully classified by ResNet50), (R) ALA attacked images that are incorrectly classified by the same ResNet50 network, with imperceptible lightness shift. The three line charts showcase the lightness value shift function generated by using our attack method. 
    %
    }
    \label{fig:effect}
    \vspace{-15pt}
\end{figure}

\section{Introduction}
Deep neural networks (DNNs) are widely used in computer vision tasks.
%
%
However, there are many attack approaches that can do harm to DNNs, especially those called \emph{adversarial attacks} \cite{szegedy2013intriguing}, which design deceiving inputs to mislead the DNNs into making wrong predictions. This kind of attack is widely used in various domains.
%

The well-known approach for generating adversarial examples is crafting $L_p$ norm-restricted human-imperceptible noise on clean images. However, the restricted perturbations can be defended by methods such as adversarial denoising \cite{Xie_2019_CVPR}. Furthermore, $L_p$ norm adversarial attacks lack practicality in both the digital and physical worlds. Most imperceptible attacks generate floating-point adversarial examples, which would lose their aggressivity after being saved as integers (\ie, the storage format of images in computers) \cite{bu2021taking}. It is also difficult to simulate imperceptible noises in the physical world with common carriers such as patches, and paints.



Therefore, studies on exploring \emph{non-suspicious adversarial images} that allow unrestricted but unnoticeable perturbations have been proposed in recent years. Geometric attacks \cite{kanbak2018geometric,engstrom2019exploring}, semantic attacks \cite{sharif2016accessorize, joshi2019semantic} and color attacks \cite{zhao2020adversarial,shamsabadi2020colorfool,shamsabadi2021semantically} are the three main aspects.
However, these attack methods are usually contrary to common sense, that is, creating things that seem strange in the real world.
Geometric attacks are obvious to human eyes due to the destruction of regular boundaries of objects. It may generate irregularly twisted monitors or desks, \etc. 
Semantic attacks modify the target in the images by adding scene-mismatch objects/textures, which usually makes the image become unnatural. For example, \cite{joshi2019semantic} generates adversarial face images that look quite different from the original faces and seem strange.
Compared with geometric attacks and semantic attacks, the adversarial examples generated by color attacks look more natural. 
They mainly apply uniform transformation in similar colors \cite{zhao2020adversarial} or colors of closed pixels \cite{shamsabadi2020colorfool}. 
In this way, the modified images show the potential to be non-suspicious to human eyes even though the images have large perturbations.
Nevertheless, even the adversarial examples generated by color attacks may not be natural enough to deceive human eyes. For example,
ColorFool \cite{shamsabadi2020colorfool} modifies the image color according to semantic segmentation divided regions, thus its performance is strongly related to the result of segmentation, \eg, if a sea is divided into two parts, then in the attacked image it may have two different colors.
ACE \cite{zhao2020adversarial} uses a simple filter to carry out the color attack. However, it may generate images with unusual objects (\eg, purple river, green sky).

Since the previous attacks are semantic interference and arouse suspicion, an attack method that does not easily cause semantic aberration is imperative.
There is a simple observation that the variation of lightness (even large variation) in images results in very little semantic change. It basically does not change the shape, texture, and color of the objects in the original images, \ie, it does not generate images containing objects that are contrary to common sense. It essentially just constructs a new light condition for the target scene.
In the real world, due to the variation of light intensity and the number of light sources, it is common to take images of different lightness with respect to the same scene and less likely to arouse suspicion.
Because of these advantages, the lightness attack seems a promising direction to generate non-suspicious unrestricted adversarial images. Thus we propose \textbf{A}dversarial \textbf{L}ightness \textbf{A}ttack (ALA), a novel lightness adjustment approach to generate natural adversarial images by applying and improving a differentiable filter \cite{hu2018exposure} that was originally used to adjust the image attribute in image processing. Using a filter-based attack has three advantages: \ding{182} The filter is human-understandable, which can guide the lightness attack in the real world, \ding{183} The filter is differentiable, which is time-saving than search-based attacks (\eg, ColorFool \cite{shamsabadi2020colorfool}). \ding{184} The filter is lightweight and resource-saving (only dozens of parameters).
However, the adversarial examples generated by directly using a monotonic lightness filter achieve low attack performance and low image quality. To obtain better attack performance and image quality, we propose two improvements: \ding{182} unconstrained enhancement, \ding{183} naturalness-aware regularization. The effectiveness of ALA is verified on two popular datasets for different tasks (\ie, ImageNet for image classification and Places-365 for scene recognition).

To sum up, our work has the following contributions:
\begin{itemize}[itemsep=0pt,topsep=0pt,parsep=0pt]
\item To our best knowledge, we are the first to research adversarial attacks by focusing on adjusting image lightness with a human-understandable filter which is extremely lightweight.
\item We propose a specialized unconstrained enhancement to improve the attack success rate by utilizing a non-monotonic filter and random initialization. We also propose naturalness-aware regularization to enhance the image quality of adversarial examples by adding a lightness range constraint and lightness distribution constraint. 
\item The experiments conducted on two datasets with different tasks shows the effectiveness of ALA in generating excellent attack performance and high-fidelity attack examples.
\end{itemize}


\section{Related Work}

\subsection{Restricted Adversarial Attacks}\label{sec:unrestricted_attack}
Traditional adversarial attacks mainly focus on generating adversarial examples with limited noise. Most researchers use small $L_p$ norm \cite{carlini2017towards,madry2017towards} to ensure this. Here are two typical methods: PGD and C\&W.
%
PGD \cite{madry2017towards} is based on projected gradient descent. I-FGSM \cite{kurakin2016adversarial} iteratively determines the perturbation in $L_p$ norm boundary with gradient information. PGD starts the I-FGSM attack with a random point within the $L_p$ norm boundary. Let $\mathbf{I} \in \mathds{R}^{H \times W \times 3}$ be the original image, and $\ell\in\mathds{1}^{K}$ its ground truth for a $K$-classification problem. For a target model $\mathcal{M}(\cdot)$, $\mathcal{M(\mathbf{I})}=\ell$, adversarial attacks aim to generate an adversarial image $\mathbf{\hat{I}}$ according to $\mathbf{I}$ to mislead the model $\mathcal{M}(\cdot)$, \ie{, $\mathcal{M(\mathbf{\hat{I}})}\neq\ell$}. 
Carlini and Wagner Attacks (C\&W) \cite{carlini2017towards} can be formulated as: $\min_{\delta} \Vert \mathbf{\hat{I}}-\mathbf{I} \Vert^{2}_{p} + \lambda \cdot \mathcal{L}_{\mathrm{C\&W}}(\mathbf{\hat{I}},\ell)$, where $\mathcal{L}_{\mathrm{C\&W}}(\mathbf{\hat{I}}, \ell )=\max(\mathcal{Z(\mathbf{\hat{I}})}_{\ell}-\max\{\mathcal{Z(\mathbf{\hat{I}})}_{i}:i\neq \ell \}, -\kappa)$ and $\mathbf{\hat{I}}=\frac{1}{2}(\tanh(\mathrm{arctanh}(\mathbf{I})+\delta)+1)$. Perturbation $\delta= \mathbf{\hat{I}}-\mathbf{I}$, and $\lambda$ is a constant selected by search. $\mathcal{Z(\cdot)}_{i}$ is the $i$-th class in logit of target model $\mathcal{M}(\cdot)$, and $\kappa$ controls the confidence level of misclassification.


\subsection{Unrestricted Adversarial Attacks}
There are mainly three kinds of unrestricted attacks: geometric attacks, semantic attacks, and 
color attacks. 
Specifically, geometric attacks \cite{kanbak2018geometric,engstrom2019exploring} implement affine transformation to original images, making the generated images too suspicious. For example, there exist non-image parts (\ie, black border) after rotating the image. This property is also employed to conduct adversarial blur attacks \cite{neurips20_abba,guo2021learning}. Semantic attacks don't consider keeping the semantic stable (\eg, \cite{sharif2016accessorize} using a crafted adversarial eyeglass frame to mislead a face recognition system), leading to human-suspicion.
Color attack is a feasible way to obtain non-suspicious examples for its uniformity when modifying images.
%
%
%
ColorFool \cite{shamsabadi2020colorfool} proposes a semantic-guided black-box adversarial attack.
Adversarial Color Enhancement (ACE) \cite{zhao2020adversarial} changes the color of original images by using differentiable parametric filters to piecewise modify the color curve.
FilterFool \cite{shamsabadi2021semantically} uses an FCNN to approximate traditional image processing filters (\eg, log transformation, Gamma correction, detail enhancement), which can be applied for the color attack.

The attacks that involve lightness are AVA \cite{tian2021ava}, Jedena \cite{gao2022can}, and ARA \cite{zhang2021adversarial}. AVA proposes adding vignetting to attack visual recognition. However, vignetting will reduce the image perception quality and is not common in images, which may arouse suspicion. Jedena exploits joint exposure \& noise to attack the co-saliency task, which leads to unwieldy multi-factor optimization. ARA proposes a relighting attack on the face recognition task, which is not general enough for other objects. 
%
%
Compared with the above unrestricted adversarial attacks, ALA has two obvious advantages: \ding{182} ALA, as a gradient-based attack, is more efficient than random search-based attacks. \ding{183} ALA provides a general and lightweight way to generate adversarial lightness examples with better naturalness.
\section{Adversarial Lightness Attack}
%
\begin{figure}
    \centering
    \subfigure[]{
        \includegraphics[width=0.4\linewidth]{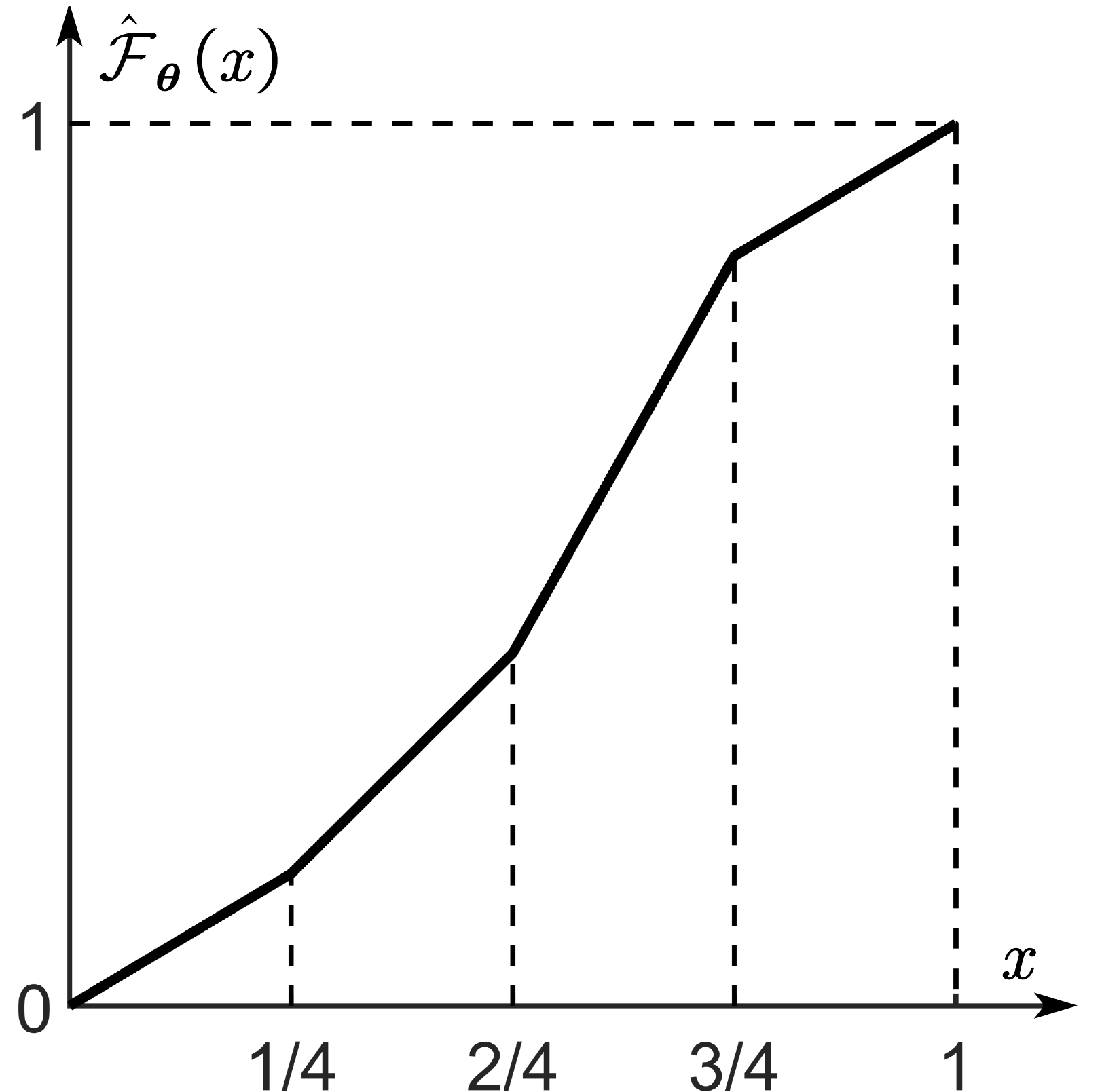}
        \label{fig:filter1}
    }
    \subfigure[]{
    	\includegraphics[width=0.4\linewidth]{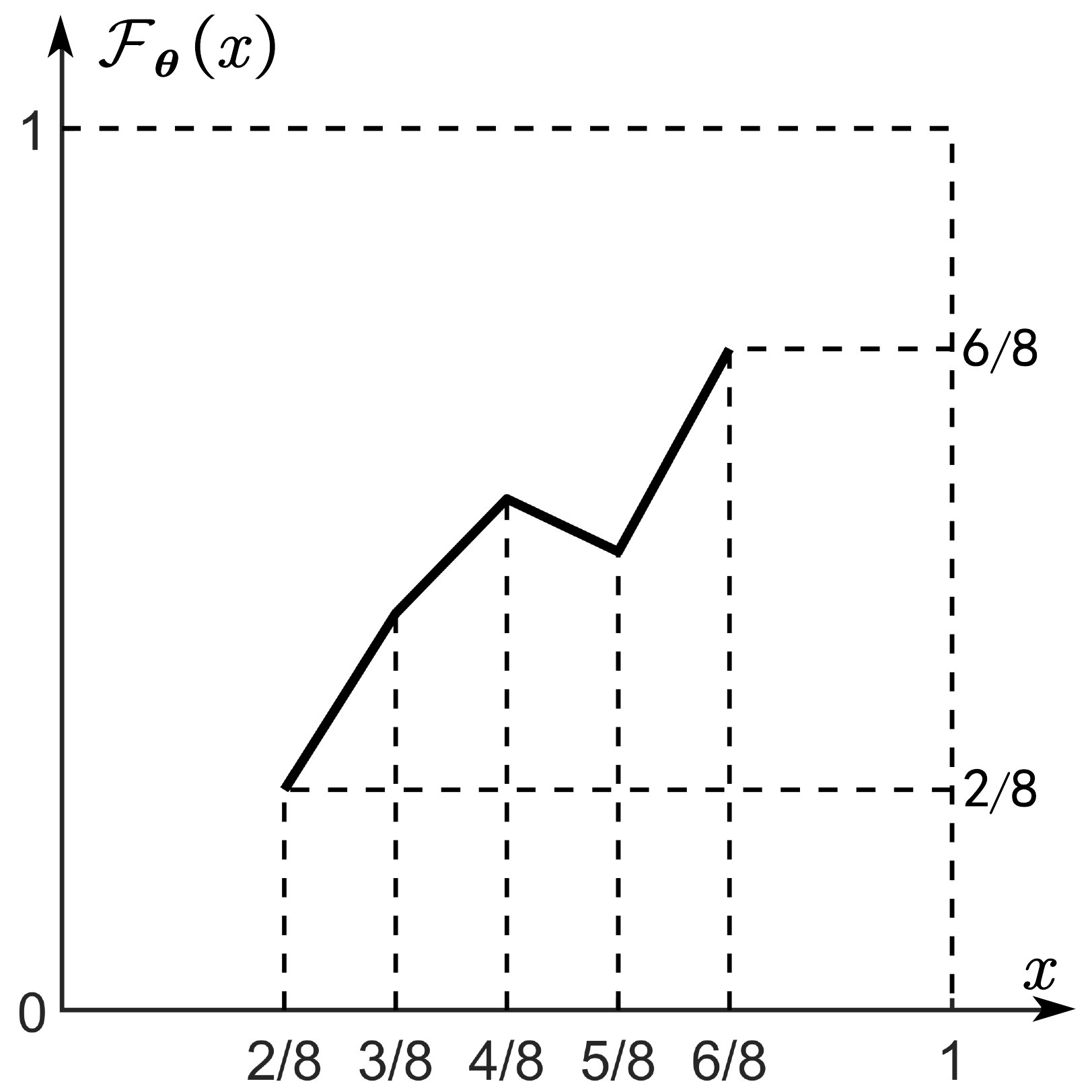}
    	\label{fig:filter2}
    }
    \caption{(a) monotonic filter $\hat{\mathcal{F}_\theta}$. (b) scene-adaptive filter $\mathcal{F}_\theta$ with the valid range from $\textbf{2/8}$ to $\textbf{6/8}$. Both filters are segmented into 4 pieces, \ie, $T=4$ in Eq. \eqref{eq:parfilter1}.}
    \label{fig:filter}
    \vspace{-15pt}
\end{figure}

\subsection{ALA via Parametric Filter}\label{sec:problem_formulation}
%
Given a clean image $\mathbf{I} \in \mathds{R}^{H \times W \times 3}$, ground truth label $l$, and a pretrained model $\mathcal{M}(\cdot)$, we aim to map $\mathbf{I}$ to a new counterpart $\hat{\mathbf{I}}$ by adjusting its surrounding lightness and make the targeted model predict incorrect category, \ie, $\mathcal{M(\mathbf{\hat{I}})} \neq \mathcal{M(\mathbf{I})}=l$.
We denote the task as the adversarial lightness attack (ALA) that is desired to output naturally relighted adversarial examples while achieving a high attack success rate.

To this end, we first convert the RGB image to LAB space and separate the lightness and color into three different channels, which ensures the changes only apply to the lightness channel. The value $\mathbf{I}_L$ in the light channel is limited to the range $[0,1]$.
Then, we design the parametric filter-based ALA to formulate the lighting variation. 
Since piece-wise linear functions are widely used in image enhancement (\eg, stretch the image contrast \cite{tsai2008contrast}, color/tone adjustments \cite{hu2018exposure}), we prefer to apply the piece-wise function for the parametric filter.
Specifically, with the targeted model to guide the optimization of lighting parameters, we can adjust $\mathbf{I}$ by
%
\begin{align} \label{eq:parfilter1}
& \hat{\mathcal{F}}_{\theta}(x) = \sum_{t=1}^T \tau_t\hat{\mathcal{F}}_{\theta_t}(x), \nonumber \\
& \text{subject to}~~\tau_t=1~\text{if}~x\in \left[\frac{t-1}{T},\frac{t}{T}\right], \text{otherwise}~\tau_t=0
\end{align}
%
where $x$ is the light value (\ie, the light channel of Lab) of one pixel in $\mathbf{I}$, and we can get a new image by replacing the light values of all pixels in $\mathbf{I}$ with the new ones generated by Eq.~\eqref{eq:parfilter1}, which is denoted as $\hat{\mathbf{I}}=\hat{\mathcal{F}}_{\theta}(\mathbf{I})$. 
Intuitively, the function Eq.~\eqref{eq:parfilter1} is a piece-wise linear function (See Fig.~\ref{fig:filter1}) with $T$ linear functions and $T$ parameters, \ie, $\theta=\{\theta_1,\ldots,\theta_T\}$. We define the $t$-th piece function as
%
\begin{align}
\hat{\mathcal{F}}_{\theta_t}(x) = \frac{T}{\sum_{t=1}^T{\theta_t}} \left(\theta_{t} \left(x-\frac{t-1}{T} \right)+\sum_{i=1}^{t-1}\frac{\theta_{i}}{T} \right).
\label{eq:piecefunc}
\end{align}
%
To allow an adversarial attack like the PGD, we tune the light-aware parameters $\theta$ with the guidance of the targeted model $\mathcal{M}$, \ie,
%
\begin{align}
&\argmax_\theta{\mathcal{L}_{\theta}(\mathcal{M}(\hat{\mathcal{F}}_{\theta}(\mathbf{I})),\ell}),\nonumber\\ 
&~\text{subject to}~ \forall \theta_t \in \theta, \theta_t > 0, \mathbf{I}_{L} \in [0,1],
\label{eq:ALA_objfunc}
\end{align}
where the valid range of value $\mathbf{I}_{L}$ represents the adjustable range of lightness values during the attack. 
Note that the restriction $\theta_t > 0$ is generally used in image enhancement to guarantee the monotonically increasing property of the parametric filter, thus maintaining the numerical magnitude relationship between pixels and achieving better image enhancement effects.
After optimizing Eq.~\eqref{eq:ALA_objfunc} via gradient descent like PGD, we can get the optimized parameter $\theta^*$ and the adversarial image can be obtained by $\mathbf{\hat{I}} = \hat{\mathcal{F}}_{\theta^{*}}(\mathbf{I})$.

In summary, using the piecewise-based filter in the lightness attack has three advantages: \ding{182} The filter is differentiable, thus showing higher efficiency than search-based methods (\eg, ColorFool \cite{shamsabadi2020colorfool}). \ding{183} The filter is more lightweight and resource-saving than other operations since the filter uses far fewer parameters (64-segment piecewise function only needs 64 parameters). \ding{184} The filter is human-understandable, which can guide the lightness attack in the real world (see Fig.~\ref{fig:physical}). 
However, simply applying $\hat{\mathcal{F}_\theta}(\cdot)$ for the lightness attack leads to low image quality and attack success rate. 

\begin{figure}
    \centering
    \subfigure[]{
        \includegraphics[width=0.3\linewidth]{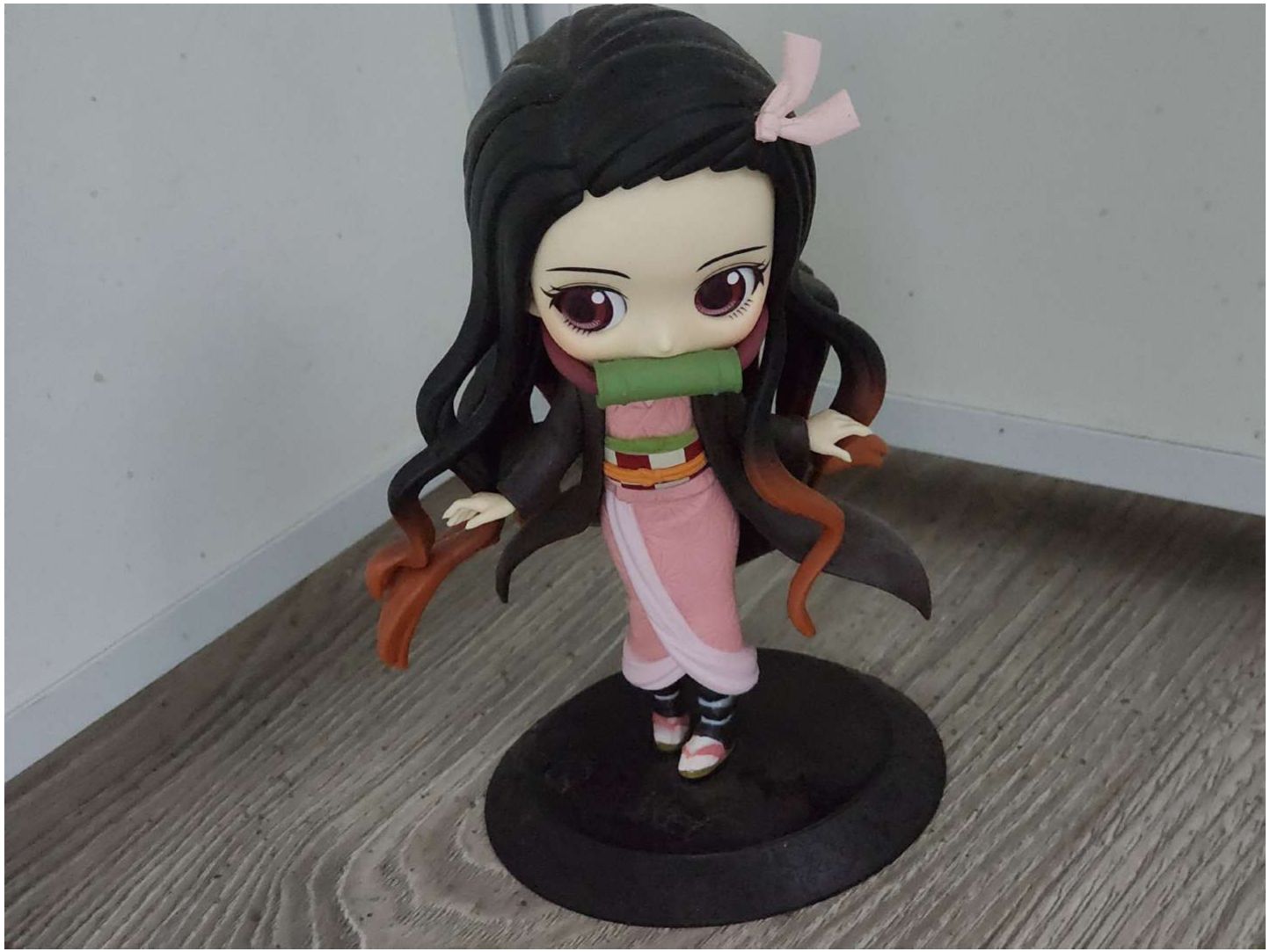}
        \label{fig:light_condition_1}
    }
    \subfigure[]{
        \includegraphics[width=0.3\linewidth]{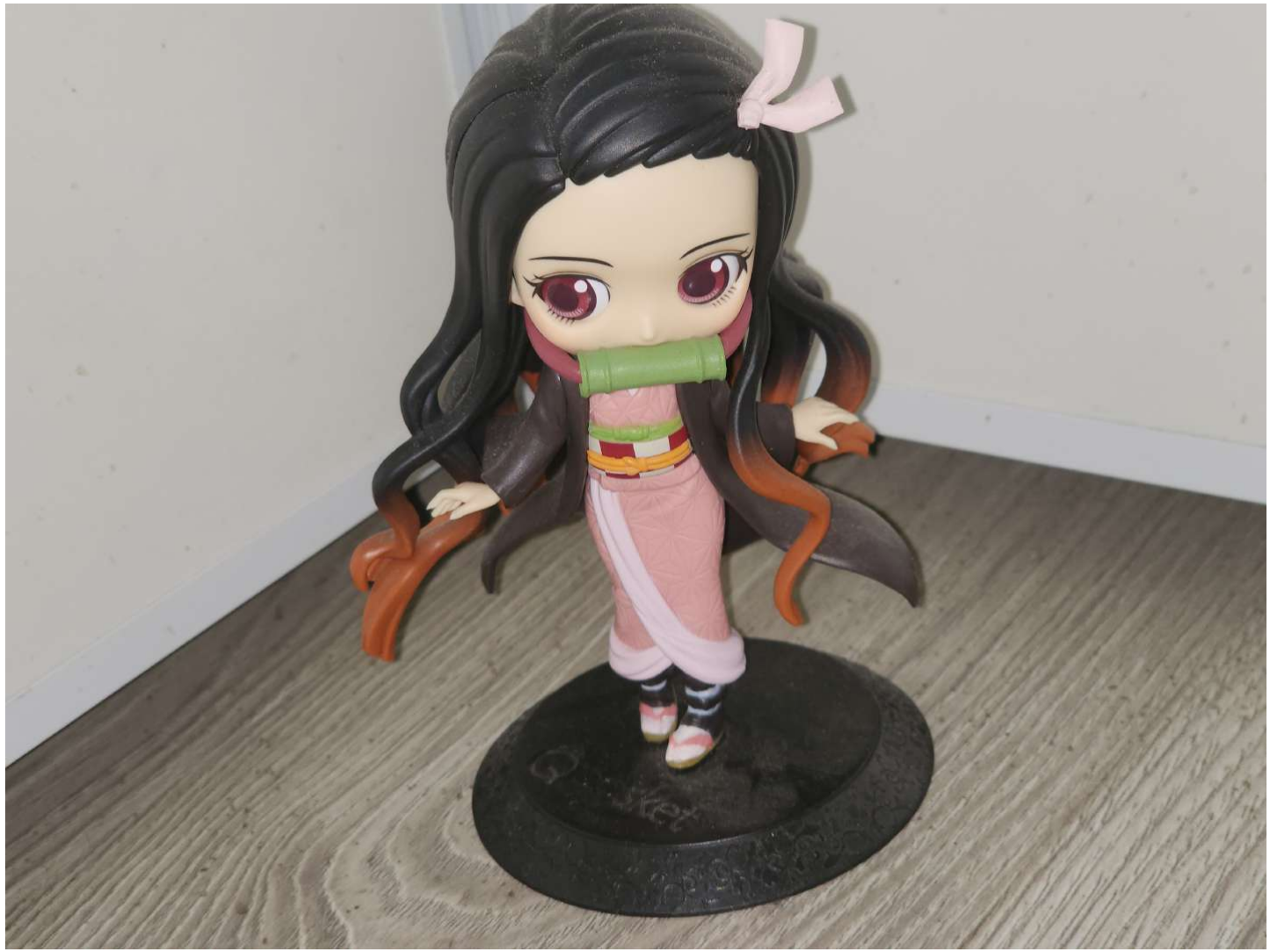}
        \label{fig:light_condition_2}
    }
    \subfigure[]{
    	\includegraphics[width=0.3\linewidth]{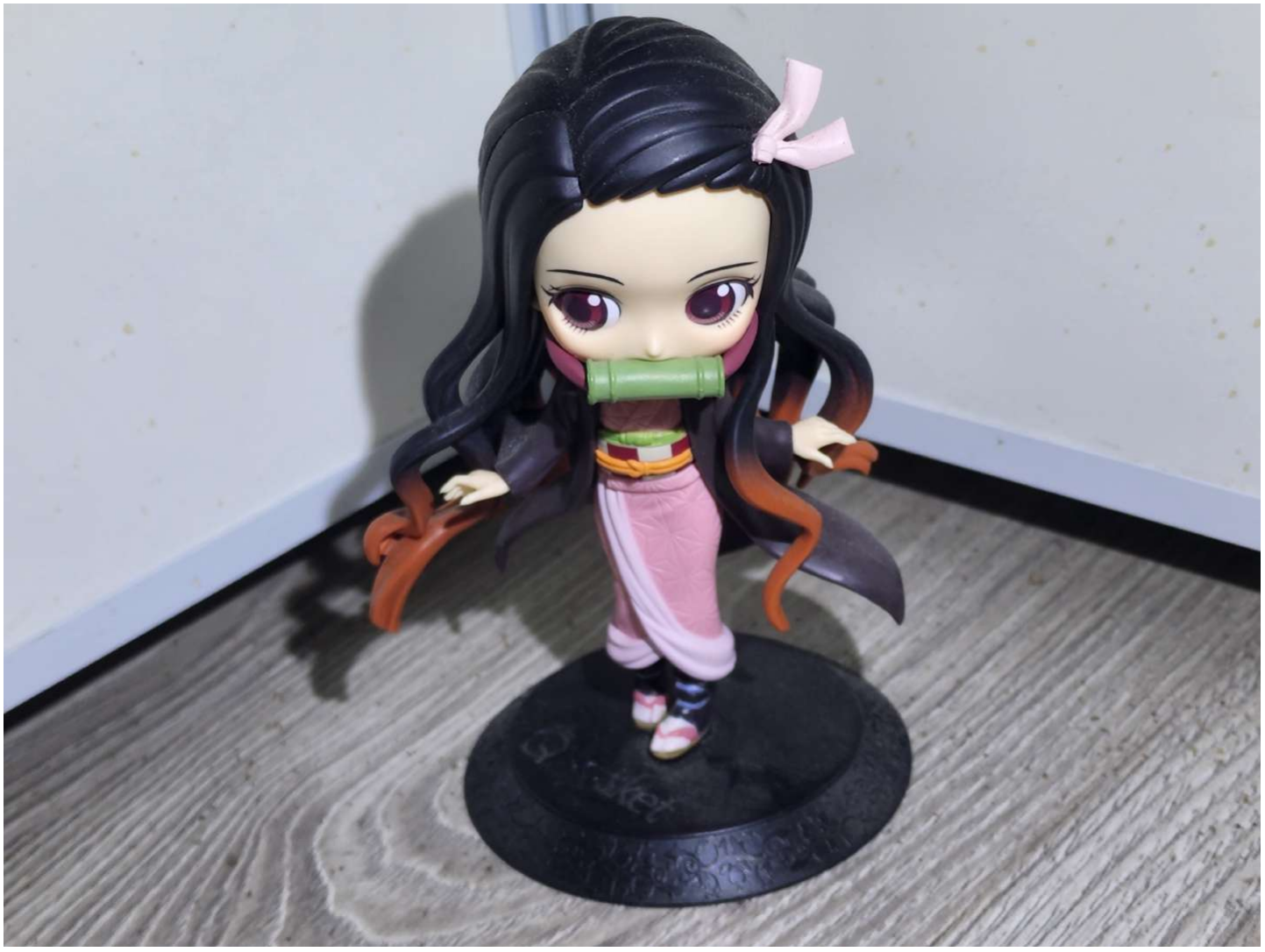}
    	\label{fig:light_condition_3}
    }
    \caption{Complex light and shade relationship in the real world. The sub-images (a), (b), (c) are under various light conditions.}
    \label{fig:light_condi}
    \vspace{-10pt}
\end{figure}

\subsection{Unconstrained Enhancement}\label{sec:unconstrained_ALA}
The attack success rate of vanilla ALA is not satisfactory (about 10\% lower than other white-box unrestricted attacks). 
According to the observation of Eq.~\eqref{eq:ALA_objfunc}, we find that the low attack success rate is mainly caused by the monotonically increasing property of the filter. It imposes strict constraints on the light and shade relationship in the image (\ie, the lighter pixels in the original image should also be lighter ones in the adversarial image), which benefits the image enhancement task but is not so necessary for attack tasks.  
Obviously, strict constraints are indispensable for image properties such as color because the incongruous color relationship that defies common sense may easily cause suspicion.
In contrast, in Fig.~\ref{fig:light_condi}, the lightness variation in the real world is extremely complex
(\eg, adding or reducing light sources to the scene of an original image can significantly influence the light and shade relationship), which cultivates people's high acceptance of the light and shade relationship. That is, in human cognition, scenes with a little unusual light and shade relationship are possible and reasonable to appear in the real world.
Since humans are perception-insensitive to lightness and there are no original images to compare with, it is feasible to make an \textbf{unconstrained enhancement}, \ie, relax restrictions of the light and shade relationship in adversarial images. 

\begin{figure}[t]
    \centering
    \includegraphics[width=0.9\linewidth]{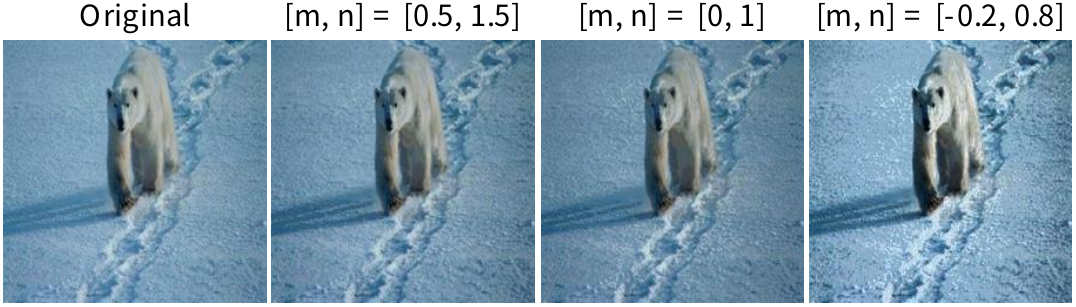}
    \caption{Adversarial examples generated by ALA with different initialization ranges.}
    \label{fig:init}
\vspace{-10pt}
\end{figure}
\begin{figure}[t]
    \centering
    \includegraphics[width=\linewidth]{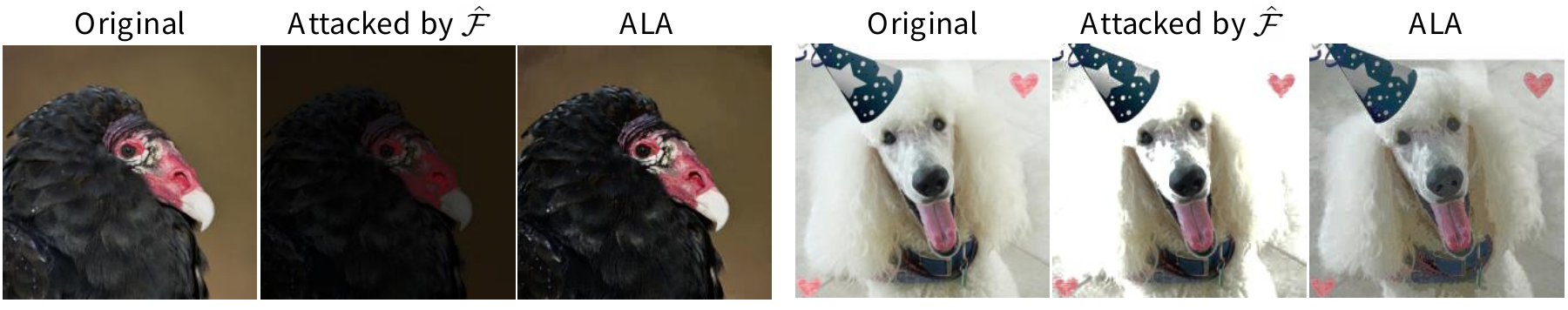}
    \caption{Adversarial images generated by different filters.}
    \label{fig:origin_filter}
    \vspace{-10pt}
\end{figure}

In specific, we exploit two strategies (\textbf{non-monotonic} and \textbf{random initialization}) to construct adversarial images with special lightness distribution that are challenging for target neural networks.
First, we lift the restriction of $\theta_{t} > 0$ in Eq.~\eqref{eq:ALA_objfunc} to get the non-strict monotonically increasing filter $\mathcal{F}_{\theta}(\cdot)$.
Second, at the beginning of the attack, we randomly initialize the parameters $\theta$ in the range of $[m, n]$ ($m<n$) instead of using parameters with one as initial values. This means the attack will not take the lightness values of the original image as the starting value for optimization. Thus the attack is more likely to generate a lightness distribution that is unfamiliar to pre-trained models. As is displayed in Fig.~\ref{fig:init}, the adversarial examples with different initialization ranges look different from the original image to varying degrees. In summary, the Eq.~\eqref{eq:ALA_objfunc} is revised as
\begin{align}
\argmax_\theta{\mathcal{L}_{\theta}(\mathcal{M}(\mathcal{F}_{\theta}(\mathbf{I})),\ell}), 
~\text{subject to}~ \mathbf{I}_{L} \in [0,1].
\label{eq:ALA_objfunc_ada_unconstrain}
\end{align}

\subsection{Naturalness-aware Regularization}\label{sec:natural_reg}
Although ALA with unconstrained enhancement significantly improves the attack success rate, the generated images look quite unnatural.
In order to improve the naturalness of images, we take a thorough review of the generated images and empirically find that the unnaturalness of images is mainly due to the common lightness-related unnatural phenomenon (\eg, overexposure, underexposure, and over-saturation), as shown in Fig.~\ref{fig:origin_filter}. These unnatural phenomena are mainly caused by abnormal fluctuations in two attributes (\ie, range and distribution) of lightness.
Thus the naturalness of images can be enhanced by directly adding lightness-related constraints to reduce the abnormal values of lightness.

In specific, we propose the naturalness-aware ALA with \textbf{lightness range constraint} and \textbf{lightness distribution constraint}.
The lightness range constraint aims to avoid extreme lightness values (\ie, overexposure and underexposure). An intuitive idea is to limit the lightness value from the valid range $[0, 1]$ to a smaller range for that it is the lightness values close to 1 and 0 cause overexposure and underexposure respectively. 
Considering that the lightness range of each image represents its scene characteristics, we propose to adjust the lightness value adaptively according to the image scene. To be specific, for an original image of which the lowest lightness value is ${\mathbf{I}^\text{min}_L}$ and the largest lightness value is ${\mathbf{I}^\text{max}_L}$, it is reasonable to set the valid range to $[{\mathbf{I}^\text{min}_L}, {\mathbf{I}^\text{max}_L}]$ (\eg, $\mathbf{I}^\text{min}_L=\frac{2}{8}$ and $\mathbf{I}^\text{max}_L=\frac{6}{8}$ in Fig.~\ref{fig:filter2}).

\begin{figure}
    \centering
    \subfigure[]{
        \includegraphics[width=0.2\linewidth]{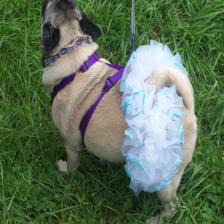}
        \label{fig:reg_ori}
    }
    \subfigure[]{
        \includegraphics[width=0.2\linewidth]{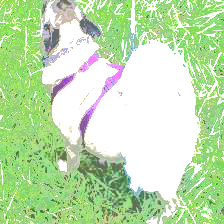}
        \label{fig:reg_not}
    }
    \subfigure[]{
    	\includegraphics[width=0.2\linewidth]{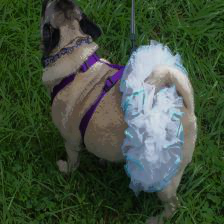}
    	\label{fig:reg}
    }
    \caption{(a) original image, (b) adversarial example filtered by $\mathcal{F}_\theta$, (c) regularized example.}
    \label{fig:reg_cmp}
    \vspace{-10pt}
\end{figure}
\begin{algorithm}
    \scriptsize
    \caption{Adversarial Lightness Attack} 
    \label{alg:ALA}
    \begin{algorithmic}[1]
    \REQUIRE{original RGB image $\mathbf{I}_{RGB}$, ground truth label $l$, target model $\mathcal{M}(\cdot)$, number of iterations $N$, learning rate $\alpha$, regularization rate $\beta$, number of segmented pieces $T$}
    \ENSURE{adversarial image $\mathbf{\hat{I}}$}
    
    \STATE Initialize $\theta^{1}\gets \mathrm{Random}(T)$
        \STATE\hspace{0.82cm}$\mathbf{I}_{LAB}\gets \mathrm{RGBtoLab}(\mathbf{I}_{RGB})$
        \STATE\hspace{0.82cm}$\mathbf{I}_{L}, \mathbf{I}_{A}, \mathbf{I}_{B}\gets \mathrm{Split}(\mathbf{I}_{LAB})$
    \FOR{$i\gets 1$ to $N$} 
        \STATE${\mathbf{I}}_{L}^{i}\gets \mathcal{F}_{\theta^{i}}({\mathbf{I}}_{L})$
        \STATE${\mathbf{I}}_{LAB}^{i}\gets \mathrm{Concatenate}({\mathbf{I}}_{L}^{i}, {\mathbf{I}}_{A}, {\mathbf{I}}_{B})$
        \STATE$\mathbf{I}_{RGB}^{i}\gets \mathrm{LabtoRGB}(\mathbf{I}_{LAB}^{i})$
        \IF{$\mathcal{M}(\mathbf{I}^{i}_{RGB}) \neq \ell$}
            \STATE$\mathbf{\hat{I}} \gets\mathbf{I}^{i}_{RGB}$
        \ENDIF
        \STATE$g \gets \nabla_{\theta}(\mathcal{L}_{\mathrm{C\&W}}(\mathbf{I}^{i}_{RGB},\ell)+\beta\cdot (-\frac{1}{T}\sum_{j=1}^T\vert\theta_j\vert))$
        \STATE$\theta^{i+1} \gets \theta^{i}-\alpha \cdot \frac{\boldsymbol{g}}{\Vert \boldsymbol{g} \Vert_{2}}$
    \ENDFOR
    \RETURN $\mathbf{\hat{I}}$ 
    \end{algorithmic}
\end{algorithm}
\label{sec:alg}
%
The lightness distribution constraint aims to avoid the tendency of pixel lightness to be similar values. Fig.~\ref{fig:reg_not} shows some adversarial images with unnatural lightness distribution. For example, many pixels change to extremely similar high lightness values, which is rare and unnatural.
It is the side effect of removing the monotonic constraint (\ie, $\theta_{t}>0$), which makes the adjacent lightness become the extremely close lightness (\ie, values of parameters $\theta$ of the pieces are close to 0). The generated images with too much ``same lightness'' region are noticeable and suspicious.

To address this problem, we need to keep the pixels of similar lightness in the original images to be different after the attack. 
Since the piecewise mapping function is represented by a set of parameters $\theta$, imposing penalties on parameters $\theta_t$ that are close to 0 can effectively avoid similar lightness values in adversarial images.
Thus we propose the lightness distribution constraint $\mathcal{L}_{\mathrm{R}}=\frac{1}{T}\sum_{t=1}^T\vert\theta_t\vert$ and the objective function Eq.~\eqref{eq:ALA_objfunc_ada_unconstrain} can be further reformulated as:
\begin{align}
&\argmax_\theta \Big({\mathcal{L}_{\theta}(\mathcal{M}(\mathcal{F}_{\theta}(\mathbf{I})),\ell})+\frac{1}{T}\sum_{t=1}^T\vert\theta_t\vert \Big),\nonumber\\
&~\text{subject to}~ \mathbf{I}_{L} \in [\mathbf{I}^\text{min}_L,\mathbf{I}^\text{max}_L].
\label{eq:ALA_objfunc_ada_reg}
\end{align}

\subsection{Algorithm of ALA}
In Algorithm \ref{alg:ALA}, we specifically show the concrete procedure of \textbf{A}dversarial \textbf{L}ightness \textbf{A}ttack (ALA). We first randomly generate $T$ values in the range of $[m, n]$ as the initial parameters $\theta^{1}$ (\ie, the first iteration of $\{\theta_{1},\theta_{2},\cdots,\theta_{T}\}$) of the mapping filter $\mathcal{F}_\theta$ in line 1, and transform the original image $\mathbf{I}_{RGB}$ to LAB color space ${\mathbf{I}}_{LAB}$ in line 2. Then we can easily get ${\mathbf{I}}_{L}$, the lightness of $\mathbf{I}$, by splitting the LAB channels of the transformed image ${\mathbf{I}}_{LAB}$. After initialization, we start to generate the adversarial attack example iteratively. In the $i$-th step, we firstly use the filter $\mathcal{F}_{{\theta}^{i}}$ to modify the original lightness channel ${\mathbf{I}}_{L}$ and obtain the new lightness channel ${\mathbf{I}}_{L}^{i}$ in line 5. Next, in line 6, we concatenate the new $L$ channel with the original $A$ and $B$ channels to get a new image in LAB color space, ${\mathbf{I}}_{LAB}^i$. In line 7, we resume it to RGB color space image, $\mathbf{I}_{RGB}^i$. In lines 8 and 9, we update the output $\mathbf{\hat{I}}$ if the $\mathbf{I}_{RGB}^i$ successfully misleads the target model $\mathcal{M}$. In line 11, we combine the $\mathcal{L}_{\mathrm{C\&W}}$ (Sec.~\ref{sec:unrestricted_attack}) with the distribution-aware regularization $\mathcal{L}_{\mathrm{R}}$ defined in Eq. \eqref{eq:ALA_objfunc_ada_reg} as the loss to compute the gradient $\boldsymbol{g}$. The $\beta$ is a hyper-parameter to control the weight of regularization. Finally, we use gradient $\boldsymbol{g}$ with learning rate $\alpha$ to update the new parameter ${\theta}^{i+1}$ of the filter $\mathcal{F}_\theta$.

\section{Experiment} \label{sec:exp}
\subsection{Experiment Setup}\label{sec:setup}
\noindent\textbf{Datasets.} We carry out our experiments on two datasets for different tasks respectively. For image classification, as ImageNet \cite{5206848} has 1,000 classes, we randomly choose 3 images per class to make up our dataset of 3,000 images with the size of 224 $\times$ 224. We perform the experiment for scene recognition on 1,095 images (randomly selected 3 images per class from Places-365 \cite{zhou2017places}, which has 365 classes) with the size of 256 $\times$ 256.

\noindent\textbf{Target model.} We choose four networks: ResNet50 \cite{he2016deep}, VGG19 \cite{simonyan2014very}, DenseNet121 \cite{huang2017densely}, and MobileNet-v2 \cite{sandler2018mobilenetv2}, for image classification. For scene recognition, we choose MobileNet-v2, ResNet50, and DenseNet161 as the target model.

\noindent\textbf{Metrics.} To evaluate the naturalness of images, we assess the images from two perspectives. To assess the human-perceptual similarity of the adversarial examples with original images, we use learned perceptual image patch similarity (LPIPS). We also use two non-reference image quality assessment methods, perception-based image quality evaluator (PIQE) and natural image quality evaluator (NIQE), to partly quantify the quality of the adversarial examples. All three metrics are the lower the better. 

\begin{table*}[!ht]
\centering
\caption{Comparison of seven attack baselines and our method on ImageNet. It shows the results of adversarial
attacks on four normally trained models: ResNet50, DenseNet121, VGG19 and MobileNet-v2. The first column displays the attack success rates (\textbf{Succ Rate}), and the last three columns are image quality metrics LPIPS score, PIQE score, and NIQE score, where we use red, yellow, and blue to mark the first, second, and third highest image quality for unrestricted attack methods.}
\resizebox{\linewidth}{!}{
\begin{tabular}{l|cccc|cccc|cccc|cccc}
\toprule 
Target Model & \multicolumn{4}{c}{ResNet50} & \multicolumn{4}{|c}{DenseNet121} & \multicolumn{4}{|c}{VGG19} & \multicolumn{4}{|c}{MobileNet-v2}\tabularnewline
\midrule 
Metrics & Succ Rate$\uparrow$ & LPIPS$\downarrow$ & PIQE$\downarrow$ & NIQE$\downarrow$ & Succ Rate$\uparrow$ & LPIPS$\downarrow$ & PIQE$\downarrow$ & NIQE$\downarrow$ & Succ Rate$\uparrow$ & LPIPS$\downarrow$ & PIQE$\downarrow$ & NIQE$\downarrow$ & Succ Rate$\uparrow$ & LPIPS$\downarrow$ & PIQE$\downarrow$ & NIQE$\downarrow$\tabularnewline
\midrule 
PGD & 92.87\% & 0.005 & 8.059 & 47.485  & 95.39\% & 0.005 & 8.047 & 47.628 & 94.69\% & 0.005 & 8.098 & 47.432 & 98.69\%  & 0.005 & 8.032 & 47.364\tabularnewline
C\&W & 100.00\% & 0.005 & 10.449 & 47.783 & 100.00\% & 0.006 & 10.715 & 47.769 & 100.00\% & 0.004 & 10.824 & 47.775 & 100.00\% & 0.004 & 10.086 & 48.022\tabularnewline
MIM & 95.71\% & 0.051 & 4.302 & 30.443 & 99.86\% & 0.053 & 4.667 & 32.380 & 98.57\% & 0.047 & 4.772 & 32.363 & 99.29\% & 0.049 & 4.474 & 31.469\tabularnewline
\midrule
ColorFool & 90.64\% & 0.208 & 13.417 & 44.577 & 84.11\% & 0.229 & 13.834 & 44.085 & 91.35\% & 0.205 & 13.520 & 44.674 & 91.98\% & 0.185 & 13.301 & 44.885\tabularnewline
ACE & 96.80\% & 0.297  & \cellcolor{tab_blue}12.771 & 41.603 & 95.72\% & 0.288 & \cellcolor{tab_blue}12.870 & \cellcolor{tab_blue}41.389  & 98.92\% & 0.295 & \cellcolor{tab_blue}12.698 & \cellcolor{tab_blue}41.073 & 98.34\% & 0.297 & \cellcolor{tab_blue}12.734  & 40.904\tabularnewline
EdgeFool & 99.27\% & \cellcolor{tab_blue}0.127 & \cellcolor{tab_yellow}12.185  & \cellcolor{tab_yellow}38.663 & 98.54\%  & \cellcolor{tab_blue}0.126  & \cellcolor{tab_yellow}12.134  & \cellcolor{tab_yellow}38.763  & 99.16\%  & \cellcolor{tab_blue}0.127  & \cellcolor{tab_yellow}12.097  & \cellcolor{tab_yellow}38.267  & 99.39\% & \cellcolor{tab_blue}0.125 & \cellcolor{tab_yellow}12.101  & \cellcolor{tab_yellow}38.668\tabularnewline
FilterFool & 100.00\% & \cellcolor{tab_red}0.111 & 16.427  & \cellcolor{tab_blue}39.688 & 100.00\% & \cellcolor{tab_red}0.111 &  16.754 & 42.344 & 100.00\% & \cellcolor{tab_red}0.109 & 16.065 & 42.071 & 100.00\% & \cellcolor{tab_yellow}0.112 & 16.021  & \cellcolor{tab_blue}38.802\tabularnewline
\textbf{ALA (Ours)} & 97.53\% & \cellcolor{tab_yellow}0.124 & \cellcolor{tab_red}10.406  & \cellcolor{tab_red}28.636 & 96.19\% & \cellcolor{tab_yellow}0.125  & \cellcolor{tab_red}10.547 & \cellcolor{tab_red}28.657  & 98.97\% & \cellcolor{tab_yellow}0.110 & \cellcolor{tab_red}9.961  & \cellcolor{tab_red}{28.472} & 99.14\% & \cellcolor{tab_red}0.109 & \cellcolor{tab_red}10.093  & \cellcolor{tab_red}28.938\tabularnewline
\bottomrule
\end{tabular}
}
\label{tab:quality}
\end{table*}
%
\noindent\textbf{Baseline methods.} We choose three $L_p$ norm-restricted adversarial attacks and four unrestricted attacks as baselines. For restricted methods, we use PGD \cite{madry2017towards} with 10 iterations and $\epsilon=2/255$, C\&W attack \cite{carlini2017towards} with 5 $\times$ 200 iterations and confidence $\kappa=20$, and MIM with $\epsilon=4/255$ for similar image quality with ALA. MIM \cite{dong2018boosting} combines the momentum term with I-FGSM for a better transfer attack success rate. For unrestricted attack methods, we follow the experiment settings of ACE \cite{zhao2020adversarial}, ColorFool \cite{shamsabadi2020colorfool}, EdgeFool \cite{shamsabadi2020edgefool}, and apply FilterFool \cite{shamsabadi2021semantically} with 1,500 iterations and stopping threshold $\tau = 0.006$. Although the settings for FilterFool are not the same as in the paper of 3,000 iterations and $\tau = 0.004$, the results are similar.

\noindent\textbf{Implementation details.} We divide the valid lightness into 64 pieces, \ie{, $T=64$ in Eq.~\eqref{eq:parfilter1}}. We set the learning rate $\alpha=0.5$ and the number of iterations $N=100$. As mentioned in Sec \ref{sec:alg}, we use $\mathcal{L}_{\mathrm{C\&W}}$ within $\kappa = 0.2$ and the regularization $\mathcal{L}_{\mathrm{R}}$ with $\beta = 0.3$ as the loss function. And the random initialization range is set as $[m, n]=[-0.2, 0.8]$. All the experiments were run on Pytorch 1.8, and CUDA 11.1 with an NVIDIA GeForce RTX 2070 SUPER GPU.

\begin{figure*}[ht]
    \centering
    \includegraphics[width=0.8\linewidth]{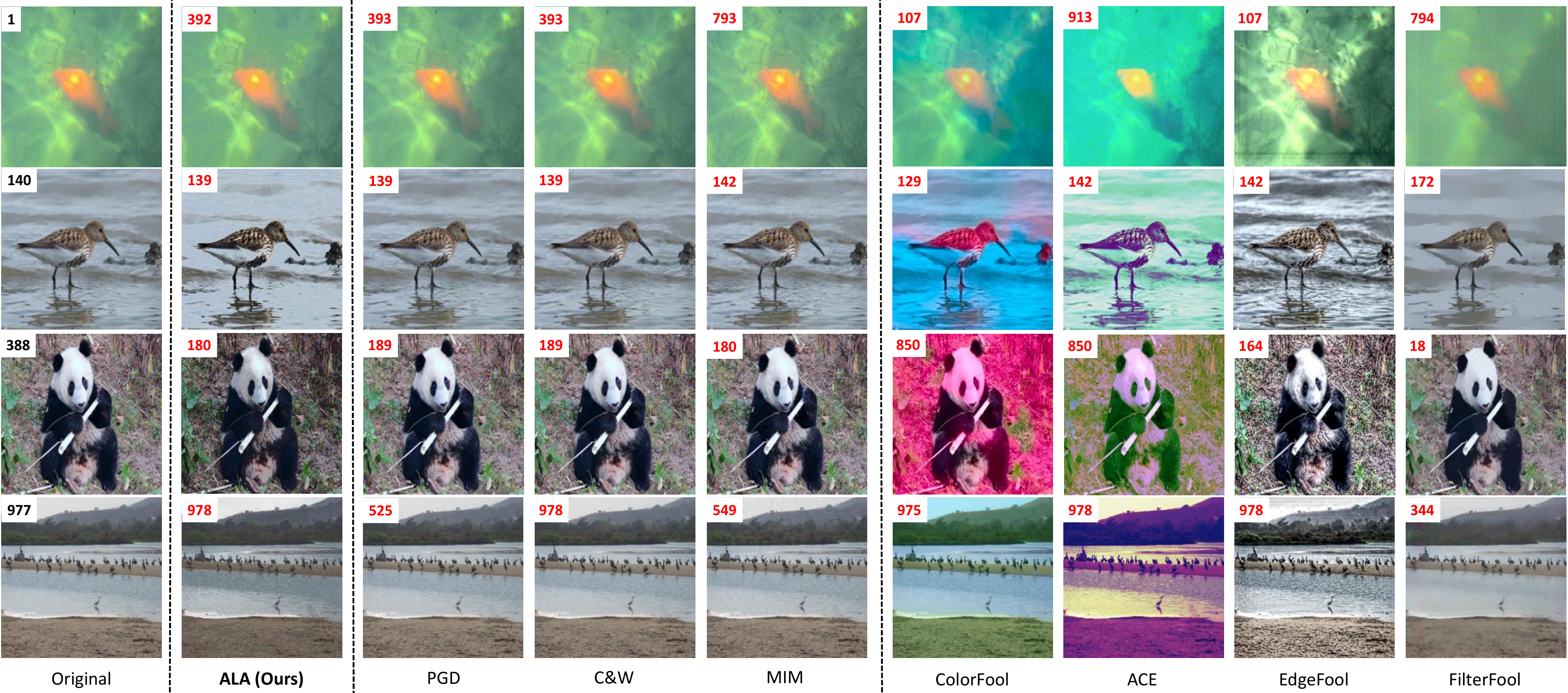}
    \caption{Adversarial examples. The top left corner shows the predicted result (ImageNet
    index) by MobileNet-v2.}
    \label{fig:res_cmp}
\end{figure*}

\begin{table*}[t]
\centering
\caption{Comparison of transferability on baselines and ALA. It shows the success rates of transfer attacks on four standard trained models: ResNet50 (\textbf{ResNet}), DenseNet121 (\textbf{DenseNet}), VGG19 (\textbf{VGG}), and MobileNet-v2 (\textbf{MobileNet}).}
\resizebox{0.85\linewidth}{!}{
\begin{tabular}{l|ccc|ccc|ccc|ccc}
\toprule 
Target Model & \multicolumn{3}{|c}{ResNet50} & \multicolumn{3}{|c}{DenseNet121} & \multicolumn{3}{|c}{VGG19} & \multicolumn{3}{|c}{MobileNet-v2}\tabularnewline
\midrule 
Attacked Model & DenseNet & VGG & MobileNet & ResNet & VGG & MobileNet & ResNet & DenseNet & MobileNet & ResNet & DenseNet & VGG\tabularnewline
\midrule 
PGD & 6.39\% & 4.77\% & 6.81\% & 6.90\% & 4.82\% & 6.00\% & 4.29\% & 4.37\% & 6.51\% & 4.16\% & 4.37\% & 5.11\%\tabularnewline
C\&W & 10.06\%  & 9.73\%  & 10.19\%  & 15.30\%  & 11.60\%  & 12.66\%  & 6.26\%  & 5.74\%  & 8.88\%  & 7.35\%  & 6.39\%  & 8.75\% \tabularnewline
MIM & 35.17\%  & 29.83\%  & 31.97\%  & 43.03\%  & 37.30\%  & 40.60\%  & 24.07\% & 24.73\%  & 31.27\%  & 25.54\%  & 26.52\%  & 33.61\% \tabularnewline
\midrule
ColorFool & 19.42\%  & 31.45\%  & 36.91\%  & 32.89\%  & 36.61\%  & 41.50\%  & 21.43\%  & 15.33\%  & 30.86\%  & 18.23\% & 13.82\%  & 23.83\% \tabularnewline
ACE & 41.00\% & 61.67\% & 58.04\% & 51.16\% & 60.29\% & 57.79\% & 46.19\% & 37.33\% & 54.87\% & 46.73\% & 36.58\%  &58.77\% \tabularnewline
EdgeFool & 23.65\% & 34.05\%  & 32.63\%  & 25.26\%  & 32.43\%  & 31.06\%  & 23.71\%  & 20.36\%  & 29.25\%  & 23.30\% & 20.73\%  & 30.12\% \tabularnewline
FilterFool & 28.82\% & 43.78\%  & 37.82\%  & 28.32\%
 & 42.80\%
 & 37.07\%
 & 24.12\%  & 22.99\%  & 34.44\%  & 24.17\% & 23.23\%  & 41.18\% \tabularnewline
\textbf{ALA (Ours)} & 28.63\%  & 44.08\%  & 43.42\%  & 33.17\%  & 44.62\%  & 43.77\%  & 24.67\%  & 20.40\%  & 33.84\%  & 23.57\% & 19.37\%  & 34.10\% \tabularnewline
\bottomrule
\end{tabular}
}
\label{tab:trans}
\end{table*}

\vspace{-5pt}
\subsection{Image Classification}\label{sec:image_class}
We compare the effectiveness of ALA with the baselines. In Table \ref{tab:quality}, most of the attack methods achieve very high success rates except for the black-box method ColorFool. Restricted methods PGD, C\&W, and MIM exploit the gradient of every pixel, thus they can successfully attack almost all images. EdgeFool and FilterFool obtain high success rates, but they cost much more training resources (\eg, 5,000 and 1,500 iterations with both 10.96 GFLOPs) than ACE and ALA (\eg, both are 100 iterations with 8.24 GFLOPs). 

Next, we compare the naturalness of the adversarial examples generated by different attacks. From Table \ref{tab:quality} we can see that the restricted methods obtain quite high LPIPS and PIQE scores. This is because LPIPS assesses the difference between attacked examples and original images, and PIQE mainly uses a special standard deviation to calculate the score, both of them are hardly influenced by restricted tiny perturbations. Even though, ALA obtains better PIQE scores than C\&W on more than half of models. For MIM, as we mentioned in Sec.~\ref{sec:setup}, we use the MIM with $\epsilon=4/255$ to generate examples having similar image quality with ALA. In this setting, the examples generated by MIM have worse NIQE and transferability than ALA. Regarding unrestricted attacks, the proposed ALA reaches the best performance of LPIPS and PIQE in almost all cases. FilterFool sometimes gets better LPIPS scores, but it cost more than ten times hours to train compared to ALA. Significantly, ALA clearly obtains the best NIQE performance in all cases.

In Figure \ref{fig:res_cmp}, we display the original inputs and the adversarial examples generated by baselines and ALA. The restricted examples of PGD and C\&W look highly similar to the original images, as their noises are too small to be noticed by humans, while the noise of MIM can be found when observing the image carefully. As for unrestricted attacks, ColorFool generates unnatural images, \eg, the bird and water in the top picture are of unnatural mixed colors. ACE and FilterFool look obviously have been processed, and EdgeFool looks unnatural around the outlines of the two animals. Compared with these unrestricted attack methods, ALA looks just like the same scene in different light conditions, \eg, decreasing the light intensity in the bottom image. Besides, we use 100 ResNet50 adversarial examples of each unrestricted attack method and their corresponding clean images to calculate the maximum mean discrepancy (MMD), which is often used in transfer learning for measuring the similarity between two different distributions. The lower MMD means the two distributions are more similar. ALA obtains the lowest score of 0.063, the following methods are EdgeFool (0.075), ColorFool (0.11), ACE (0.298), and FilterFool (0.690). The result shows that in unrestricted adversarial attacks, the adversarial images generated by ALA are in the domain that is most similar to the domain of natural (\ie, clean) images. Furthermore, the score of ALA on MMD is even comparable to restricted attacks (MMD of PGD is 0.0439), which fully demonstrates that the ALA is naturalness-aware. This is a very surprising result since the noise generated by PGD is minor and the adjustment by ALA is much larger.


Then we compare the transferability of different attacks in Table \ref{tab:trans}. Transfer attack means using the adversarial examples generated by the target model to attack other models. As shown in the first two rows of Table \ref{tab:trans}, although human-imperceptible restricted methods can achieve high success rates in target models, their transferability is too weak. Though ALA doesn't focus on transferability, within similar image quality, it still gains a comparable performance with MIM, which is specially designed for transfer attack. Among the five unrestricted attacks, ACE always performs best for transferability. The best two methods in the rest are ALA and FilterFool. It is noteworthy that, ACE wins the transferability at the price of image quality, and ALA obtains no worse transferability than FilterFool with much better image quality and much fewer training resources (ALA spends about 5 hours, while FilterFool spends more than 100 hours). 
All in all, among the unrestricted attacks, ALA obtains the best performance of image quality with considerable transferability and few training resources, which fully reflects the practicality of ALA to be applied in attack scenarios.

\begin{table*}[!ht]
\centering
\caption{Comparison of attack performance on Places-365. It shows the success rates (\%) of adversarial attacks on ResNet50 (\textbf{ResNet}), DenseNet161 (\textbf{DenseNet}), and MobileNet-v2 (\textbf{MobileNet}), in first three columns. In last three columns, it shows scores of three image quality metrics. We use red, yellow, and blue to mark the first, second, and third performances of unrestricted attack methods.}
\resizebox{0.9\linewidth}{!}{
\begin{tabular}{l|cccccc|cccccc|cccccc}
\toprule 
Target Model & \multicolumn{6}{c}{ResNet50} & \multicolumn{6}{|c}{DenseNet161} & \multicolumn{6}{|c}{MobileNet-v2}\tabularnewline
\midrule 
Model\&Metrics & ResNet & DenseNet & MobileNet & LPIPS$\downarrow$ & PIQE$\downarrow$ & NIQE$\downarrow$ & ResNet & DenseNet & MobileNet & LPIPS$\downarrow$ & PIQE$\downarrow$ & NIQE$\downarrow$ & ResNet & DenseNet & MobileNet & LPIPS$\downarrow$ & PIQE$\downarrow$ & NIQE$\downarrow$\tabularnewline
\midrule
PGD & 78.42 & 9.58 & 6.81 & 0.005 & 7.071 & 44.462 & 13.20 & 79.38 & 5.41 & 0.005 & 7.011 & 44.430 & 4.51 & 2.76 & 92.50 & 0.005  & 7.018 & 44.135\tabularnewline
C\&W & 100.00 & 39.77 & 28.62 & 0.034 & 9.149 & 45.580 & 46.54 & 100.00 & 28.80 & 0.036  & 8.948 & 45.863 & 17.55 & 15.10 & 100.00 & 0.037 & 9.699 & 44.135\tabularnewline
MIM & 91.30 & 49.83 & 45.20 & 0.098 & 5.569 & 29.667 & 57.00 & 92.37 & 45.54 & 0.106  & 5.488 & 29.211 & 27.53 & 24.02 & 97.90 & 0.085 & 5.943 & 28.747\tabularnewline
\midrule
ColorFool & 90.50 & 25.32 & 37.35 & 0.156 & \cellcolor{tab_blue}11.661 & 41.426 & 33.98 & 90.91 & 41.88 & 0.170 & \cellcolor{tab_blue}11.628 & 41.785 & 22.38 & 18.18 & 95.11 & 0.137 & \cellcolor{tab_blue}11.581  & 41.896\tabularnewline
ACE & \cellcolor{tab_blue}95.81 & \cellcolor{tab_red}60.39 & \cellcolor{tab_red}70.33 & 0.315  & \cellcolor{tab_yellow}11.230 & \cellcolor{tab_blue}39.597 & \cellcolor{tab_red}59.26 & \cellcolor{tab_blue}95.29 & \cellcolor{tab_red}65.10 & 0.310  & \cellcolor{tab_yellow}11.204 & \cellcolor{tab_blue}39.395 & \cellcolor{tab_red}53.14 & \cellcolor{tab_red}49.84 & \cellcolor{tab_blue}97.91 & 0.313 & \cellcolor{tab_yellow}11.262 & \cellcolor{tab_blue}38.961\tabularnewline
EdgeFool & \cellcolor{tab_red}99.52 & \cellcolor{tab_blue}27.76 & \cellcolor{tab_blue}43.98 & \cellcolor{tab_yellow}0.126  & 13.036  & \cellcolor{tab_yellow}33.356 & 32.85  & \cellcolor{tab_red}99.84  & \cellcolor{tab_blue}43.28 & \cellcolor{tab_blue}0.127 & 13.125 & \cellcolor{tab_yellow}33.017 & \cellcolor{tab_blue}29.47 & \cellcolor{tab_blue}25.49 & \cellcolor{tab_red}99.13 & \cellcolor{tab_blue}0.125 & 13.164  & \cellcolor{tab_yellow}33.148\tabularnewline
FilterFool & 89.74 & 23.37 & 27.02 & \cellcolor{tab_red}0.086  & 18.070 & 46.872 & \cellcolor{tab_blue}41.02  & 91.42  & 23.52 & \cellcolor{tab_red}0.083 & 19.277 & 48.277 & 28.40 & 24.17 & 92.94 & \cellcolor{tab_red}0.086 & 17.721  & 47.585\tabularnewline
\textbf{ALA(Ours)} & \cellcolor{tab_yellow}98.87 & \cellcolor{tab_yellow}43.02 & \cellcolor{tab_yellow}54.45 & \cellcolor{tab_blue}0.131 & \cellcolor{tab_red}11.193 & \cellcolor{tab_red}27.599 & \cellcolor{tab_yellow}44.15 & \cellcolor{tab_yellow}96.43 & \cellcolor{tab_yellow}45.90 & \cellcolor{tab_yellow}0.127 & \cellcolor{tab_red}11.113 & \cellcolor{tab_red}27.675 & \cellcolor{tab_yellow}31.24 & \cellcolor{tab_yellow}30.52 & \cellcolor{tab_yellow}98.95 & \cellcolor{tab_yellow}0.117 & \cellcolor{tab_red}10.667 & \cellcolor{tab_red}27.872\tabularnewline
\bottomrule
\end{tabular}
}
\label{tab:places}
\end{table*}

\begin{figure*}[t]
    \centering
    \includegraphics[width=0.8\linewidth]{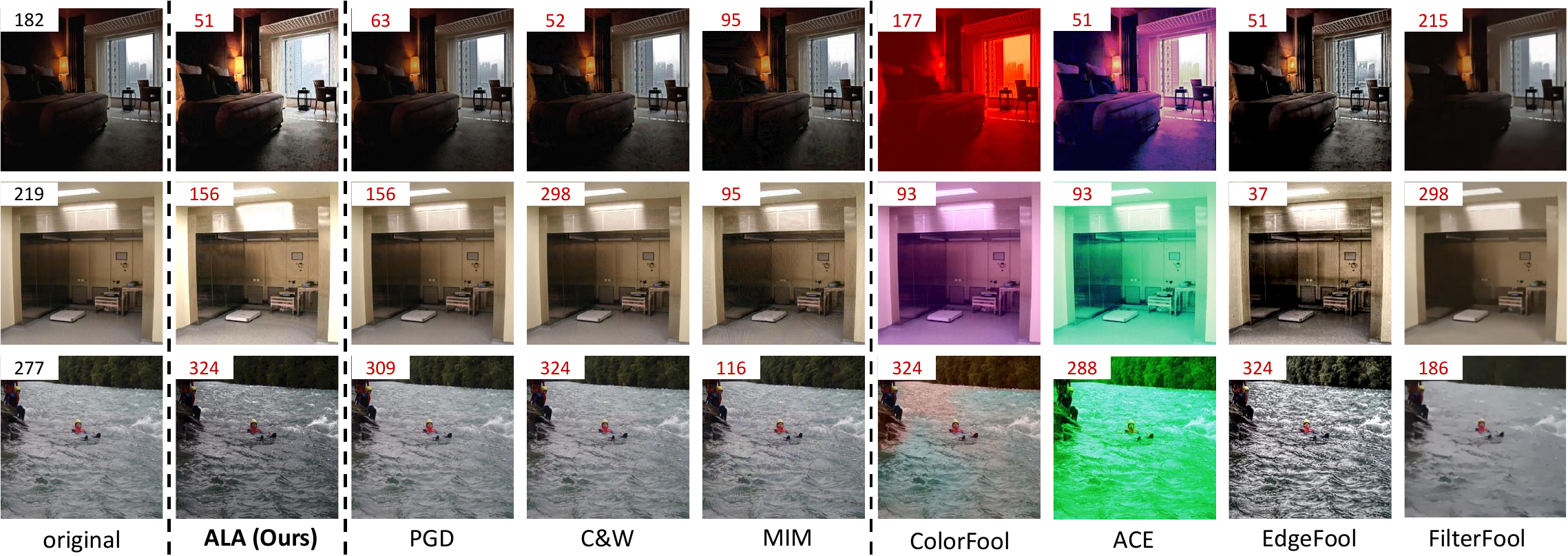}
    \caption{Adversarial examples. The top left corner shows the predicted result (Places-365
    index) by MobileNet-v2.}
    \label{fig:places_cmp}
\vspace{-15pt}
\end{figure*}
\begin{figure}
    \centering
    \subfigure[]{
        \includegraphics[width=0.22\linewidth]{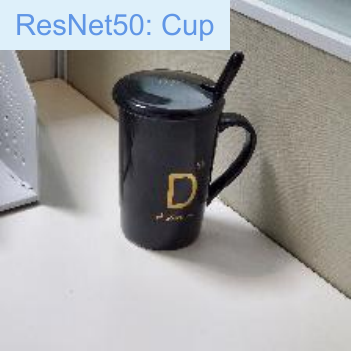}
        \label{fig:ori_504}
    }
    \subfigure[]{
        \includegraphics[width=0.22\linewidth]{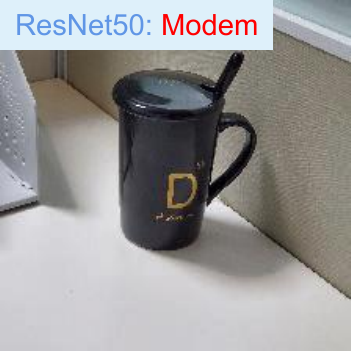}
        \label{fig:ala_633}
    }
    \subfigure[]{
        \includegraphics[width=0.22\linewidth]{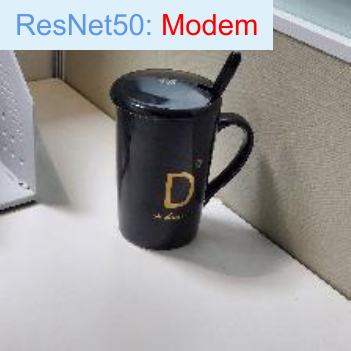}
        \label{fig:phy_633}
    }
    \caption{(a) Original image, (b) ALA image, (c) Physical image.}
    \label{fig:physical}
    \vspace{-15pt}
\end{figure}
\subsection{Scene Recognition}
Scene recognition is a challenging task, as it needs to understand the context that consists of different objects and various relationships. Thus the attack performance of different methods on the Places-365 dataset is quite different from the results of ImageNet. For example, as shown in Table \ref{tab:places}, in terms of the restricted attack methods, the attack performance of PGD is not satisfactory. It has a poor attack success rate of lower than 80\% on the target model and rarely has attack transferability. 

In the unrestricted attack methods, ALA achieves the second-best attack success rate and second-best attack transferability with the best image quality score. Especially, ALA gets the best NIQE scores among all baselines including restricted and unrestricted attack methods. Fig.~\ref{fig:places_cmp} displays the adversarial examples of baselines on Places-365. Among unrestricted attacks, the adversarial examples generated by ALA show the best naturalness and image quality.

\subsection{Real-world Attack}
As an attack method that adjusts natural phenomena (\ie, light), ALA has a certain possibility to be implemented in the physical world by manually modifying the number and angle of light sources. 
%
To simulate the lightness attack in the real world, we first take a photo of the clean image from the real world. Then we generate an adversarial image by ALA. To conveniently simulate the adversarial examples, we decrease the segment number of ALA, \ie, $T$ = 16 in Eq.~\eqref{eq:parfilter1} and don't use random initialization. Third, we take the adversarial example as the reference and construct a light condition similar to it. Finally, we take a photo from the same view as step one. In most cases, such manually-built adversarial examples in the real world can successfully fool the target model.

We carry out this experiment on 50 objects and finally get an attack success rate of 80\%. One of the attacked objects is shown in Figure~\ref{fig:physical}, we take a photo of a cup (Fig.~\ref{fig:ori_504}) in the real world and classify this image by ResNet50. Then we attack this image by ALA (Fig.~\ref{fig:ala_633}) and successfully fool the model. Next, we simulate the same light condition with the digital adversarial image generated by ALA in the real world and still take a photo of the cup. The new image (Fig.~\ref{fig:phy_633}) also mislead the ResNet50. The attack effect between Fig.~\ref{fig:ori_504} and Fig.~\ref{fig:phy_633} is mainly reflected in the fact that the shadows of the bookshelves and cups become darker in Fig.~\ref{fig:phy_633}.

\begin{table*}[t]
\centering
\caption{Ablation study. \textbf{Range}, \textbf{Dist}, \textbf{N-Mono}, and \textbf{Rand} mean lightness range constraint, lightness distribution constraint, non-monotonic, and random initialization respectively.}
\resizebox{0.8\linewidth}{!}{
\begin{tabular}{c|c|c|c|c|ccccccc}
\toprule 
Method & \textbf{Range} & \textbf{Dist} & \textbf{N-Mono} & \textbf{Rand} & ResNet50 & DenseNet121 & VGG19 & MobileNet-V2 & LPIPS$\downarrow$ & PIQE$\downarrow$ & NIQE$\downarrow$\tabularnewline
\midrule 
ALA$_0$ &  &  &  &  & 84.33\% & 11.61\% & 24.37\% & 24.10\% & 0.072 & 14.714 & 27.975\tabularnewline
\midrule 
ALA$_1$ & $\surd$ &  &  &  & 73.69\% & 7.48\% & 14.94\% & 15.23\% & 0.041 & 13.550 & 25.499\tabularnewline
\midrule 
ALA$_2$ &  & $\surd$ &  &  & 83.65\% & 10.72\% & 22.21\% & 22.69\% & 0.066 & 14.502 & 27.714\tabularnewline
\midrule 
ALA$_3$ &  &  & $\surd$ &  & 94.24\% & 15.14\% & 29.63\% & 30.61\% & 0.080 & 14.030 & 28.286\tabularnewline
\midrule 
ALA$_4$ &  &  &  & $\surd$ & 85.56\% & 16.78\% & 30.27\% & 31.92\% & 0.097 & 15.443 & 29.408\tabularnewline
\midrule 
ALA$_5$ & $\surd$ & $\surd$ &  &  & 73.96\% & 6.91\% & 13.51\% & 14.22\% & 0.039 & 13.519 & 25.554\tabularnewline
\midrule 
ALA$_6$ &  &  & $\surd$ & $\surd$ & 99.04\% & 44.80\% & 63.98\% & 64.95\% & 0.195 & 16.229 & 38.344\tabularnewline
\midrule 
ALA$_7$ & $\surd$ & $\surd$ & $\surd$ & $\surd$ & 97.53\% & 28.63\% & 44.08\% & 43.42\% & 0.124 & 10.406 & 28.636\tabularnewline
\bottomrule
\end{tabular}
}
\label{tab:ablation}
\end{table*}

\vspace{-5pt}
\subsection{Adversarial Training}
The generated adversarial examples of ALA can also be used as a kind of data augmentation to enhance the robustness of networks on unseen lightness corruptions. We take the classification task as an example to show this. We combine 6,000 images generated by our method with original images in ImageNet (12,000 images total) as the train set to fine-tune the standard trained ResNet50 model. To verify the effectiveness, we use part of the ImageNet-C \cite{hendrycks2019robustness}, which contains the ImageNet validation set with various common corruptions (\eg, brightness corruption, jpeg compression) as the test set. As our method focuses on modifying the lightness, we use the lightness corrupted images. The standard trained ResNet50 achieves 58.93\% accuracy in the test set. After fine-tuning with the new training dataset, the accuracy increases to 62.01\%, showing the effectiveness of ALA in improving the robustness of DNNs on unseen lightness corruption. 

\vspace{-5pt}
\section{Ablation Study}\label{sec:ablation}
To verify the effectiveness of the proposed lightness range constraint, lightness distribution constraint, non-monotonic, and random initialization. We conduct an ablation study on each of them and their combination. The dataset is the same as the image classification task (Sec.~\ref{sec:image_class}). ALA$_0$ is the vanilla version of ALA (implementation of filter in Sec.~\ref{sec:problem_formulation}), which is of low image quality and fair attack success rate. ALA$_7$ is the final version of ALA, which exploits all the optimization strategies and finds a trade-off between image quality and attack performance. It achieves a significantly high attack success rate while maintaining the naturalness of images.

\begin{figure}
    \centering
    \subfigure[]{
        \includegraphics[width=0.3\linewidth]{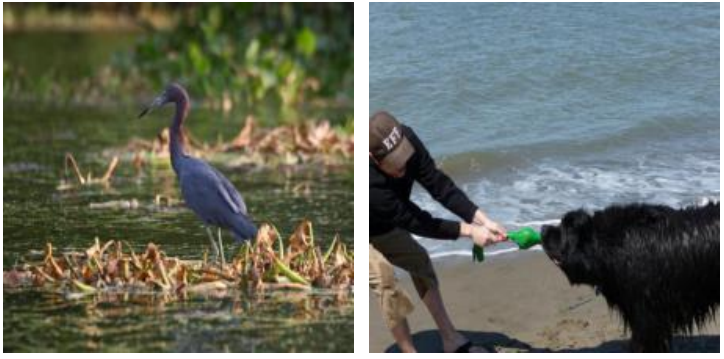}
        \label{fig:ablation_1}
    }
    \subfigure[]{
        \includegraphics[width=0.3\linewidth]{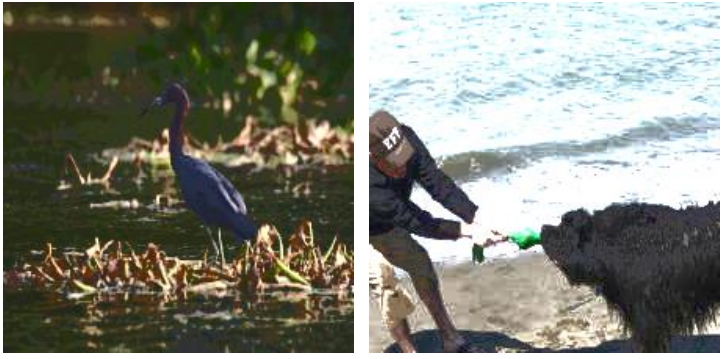}
        \label{fig:ablation_2}
    }
    \subfigure[]{
    	\includegraphics[width=0.3\linewidth]{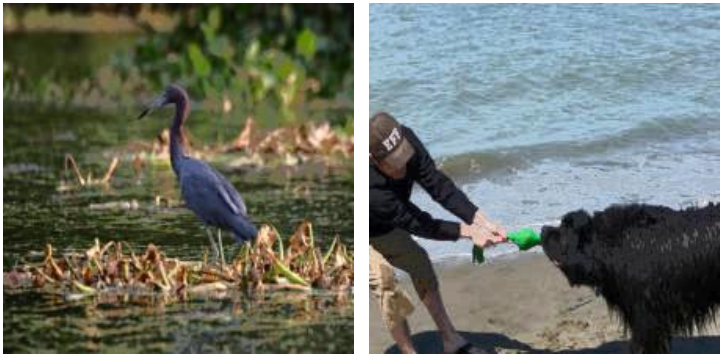}
    	\label{fig:ablation_3}
    }
    \caption{Different ALA methods examples with their original images. The sub-images (a) is the original image, (b) is generated by ALA$_6$, and (c) is generated by ALA$_7$.}
    \label{fig:ablation}
    \vspace{-10pt}
\end{figure}

\noindent\textbf{Unconstrained Enhancement.} As is mentioned in Sec.~\ref{sec:unconstrained_ALA}, we propose unconstrained enhancement (\ie, non-monotonic filter and random initialization) to improve the attack success rate. From the Table.~\ref{tab:ablation}, we can see that the ALA$_3$ and ALA$_4$, which respectively use the non-monotonic filter and random initialization, both obtain better attack success rates. Furthermore, the combined attack method, ALA$_6$, has the best attack performance. However, the images generated by ALA$_6$ have the poorest image quality on all three metrics and are human-suspicious, as is shown in Fig.~\ref{fig:ablation}. 

\noindent\textbf{Naturalness-aware Regularization.} As is mentioned in Sec.~\ref{sec:natural_reg}, we propose the naturalness-aware regularization (\ie, lightness range constraint and lightness distribution constraint) to improve the image quality. In Table.~\ref{tab:ablation}, ALA$_1$ and ALA$_2$ use the two optimizations respectively, and they both improve the image quality of adversarial examples. We combine them in ALA$_5$ to gain better image quality. However, only using the naturalness-aware regularization will result in unsatisfactory attack performance. Thus we combine the unconstrained enhancement and naturalness-aware regularization to get ALA$_7$, the comprehensive optimal ALA with both satisfactory attack performance and image quality.

\section{Discussion}
We think why the piecewise-based filter works well for the lightness attack can be attributed to three key advantages.
\ding{182} \textbf{Efficient}. The piecewise-based filter offers an efficient means of adjusting various image properties, including lightness, in a parameterized manner [1]. Therefore, it is reasonable to leverage the piecewise-based filter to construct an image property-related (e.g., lightness) attack.
\ding{183} \textbf{Unlimited}. The piecewise-based filter offers the capability to make significant adjustments to the lightness of images, which is necessary for constructing an unrestricted attack and reduces the difficulty of executing the attack.
\ding{184} \textbf{Tidy}. The piecewise-based filter optimizes the lightness value according to their different intensity levels, which avoids the adjustment of brightness to be individually or messy, ensuring the naturalness of the image.

\section{Conclusion}
We propose a novel adversarial lightness attack method ALA, which generates unrestricted examples with better naturalness than existing unrestricted adversarial attacks. 
%
%
In the future, we aim to combine other image attributes (\eg, contrast, color curve, and exposure) or degradation (\eg, noise, blur) with lightness for better performance and to explore the real-world attack more systematically. Furthermore, we are interested in exploring adversarial defense methods \cite{huang2021advfilter,qi2021archrepair,jia2022adversarial,jia2022prior} against these unrestricted attacks.

\begin{acks}
\par Geguang Pu is supported by the National Key Research and Development Program (2020AAA0107800), and Shanghai Collaborative Innovation Center of  Trusted Industry Internet Software.
This work is supported by Nanyang Technological University (NTU)-DESAY SV Research Program under Grant 2018-0980, the National Research Foundation, Singapore under its AI Singapore Programme (AISG Award No: AISG2-PhD-2021-08-022T). It is also supported by A*STAR Centre for Frontier AI Research, the National Research Foundation, Singapore, and DSO National Laboratories under the AI Singapore Programme (AISG Award No: AISG2-GC-2023-008), NRF Investigatorship No.~NRF-NRFI06-2020-0001.
\end{acks}

\balance
\bibliographystyle{ACM-Reference-Format}
\bibliography{sample-base}

\end{document}